\documentclass[letterpaper, 10 pt, conference]{ieeeconf}  

\IEEEoverridecommandlockouts                              

\usepackage{graphicx} 
\usepackage{cite}

\title{\LARGE \bf
Using Learned Predictions as Feedback to Improve Control and Communication with an Artificial Limb: Preliminary Findings
}

\author{Adam S.~R. Parker, Ann L. Edwards, and Patrick M. Pilarski$^{*}$
\thanks{All authors are with the Division of Physical Medicine and Rehabilitation and the Alberta Innovates Centre for Machine Learning, University of Alberta, Edmonton, AB, Canada, T6G 2E1. This work was support by the Alberta Innovates Centre for Machine Learning (AICML), Alberta Innovates -- Technology Futures, and the UoA Undergraduate Research Initiative.}%
\thanks{* Please direct correspondence to: {\tt\small  pilarski@ualberta.ca}}
}

\begin{document}

\maketitle
\thispagestyle{plain}
\pagestyle{plain}

\begin{abstract}
Many people suffer from the loss of a limb. Learning to get by without an arm or hand can be very challenging, and existing prostheses do not yet fulfil the needs of individuals with amputations. One promising solution is to provide greater communication between a prosthesis and its user. Towards this end, we present a simple machine learning interface to supplement the control of a robotic limb with feedback to the user about what the limb will be experiencing in the near future. A real-time prediction learner was implemented to predict impact-related electrical load experienced by a robot limb; the learning system's predictions were then communicated to the device's user to aid in their interactions with a workspace. We tested this system with five able-bodied subjects. Each subject manipulated the robot arm while receiving different forms of vibrotactile feedback regarding the arm's contact with its workspace. Our trials showed that communicable predictions could be learned quickly during human control of the robot arm. Using these predictions as a basis for feedback led to a statistically significant improvement in task performance when compared to purely reactive feedback from the device. Our study therefore contributes initial evidence that prediction learning and machine intelligence can benefit not just control, but also feedback from an artificial limb. We expect that a greater level of acceptance and ownership can be achieved if the prosthesis itself takes an active role in transmitting learned knowledge about its state and its situation of use.
\end{abstract}

\section{INTRODUCTION}

The loss of a limb can affect anyone, and prosthetic artificial limbs are often seen as a means of mitigating that loss. A person may require a prosthetic device from birth, or it could be the result of injuries sustained over the course of one's life. In both cases, but in the event of an accident especially, it can be very difficult to adapt to interacting with the world through a mechanical or electronic device \cite{Resnik2012,Peerdeman2011,Williams2011,Scheme2011,Micera2010,Parker2006,Antfolk2013,Hebert2014}. There are many prosthetics on the market that attempt to fill the needs of amputees, and many of these do an admirable job of restoring functionality and independence to the user; however, even the best prosthetics currently available have limitations \cite{Williams2011,Peerdeman2011,Resnik2012}. There are two major areas where current prosthetics begin to show the strain of insufficient technology to properly support them. The first area is a lack of feedback\cite{Micera2010,Peerdeman2011,Antfolk2013,Hebert2014}---e.g., the sense of touch---and more important to this work, lack of proprioception when using a prosthesis\cite{Williams2011,Antfolk2013}. The second area is insufficient control\cite{Micera2010,Peerdeman2011,Scheme2011,Williams2011,Parker2006}. Under most current techniques, the person who needs to control the limb has fewer control channels available to them than their device has functions\cite{Scheme2011,Parker2006,Micera2010}. This leads to some clever, but non-natural, control solutions such as routing some of the control channels to alternate locations on the user's body. A final challenge is acceptance of the prosthesis by the user\cite{Resnik2012,Williams2011,Peerdeman2011}. While many prosthetics have great clinical potential, as a result of the first two limitations, a prosthetic can be perceived by the user as insufficient, or as a reminder of the functionality that they lost and that the device simply cannot restore\cite{Resnik2012,Williams2011,Peerdeman2011}. Lack of acceptance is especially prominent in the newer myoelectric prosthetics versus the older mechanical types, despite the increased potential that electromechanical prosthetics have in overcoming the other challenges\cite{Williams2011,Scheme2011}.

Operating a device that interacts with the world is a learned motor function. As infants, we learn the way our limbs interact with our environment through general motion and play\cite{Wolpert2001,Flanagan2003,Zacks2011}. This develops the control channels and models required for us to use our bodies to sense and manipulate the world we live in \cite{Wolpert2001}. This interaction involves two parts\cite{Wolpert2001,Flanagan2003}. The first is the internal forward copy of the action---in effect, knowledge that moving specific muscles will cause a motion which results in the desired sensory feedback. There is also a reverse copy that is processed at the same time. The reverse copy starts at the desired interaction with the environment and links the required muscle action to it. In order to skillfully interact with the environment, both the forward and reverse models must be present \cite{Wolpert2001,Flanagan2003}.

\begin{figure}[t]
\vspace{0.6em}
\centering
\includegraphics[width=0.91\linewidth]{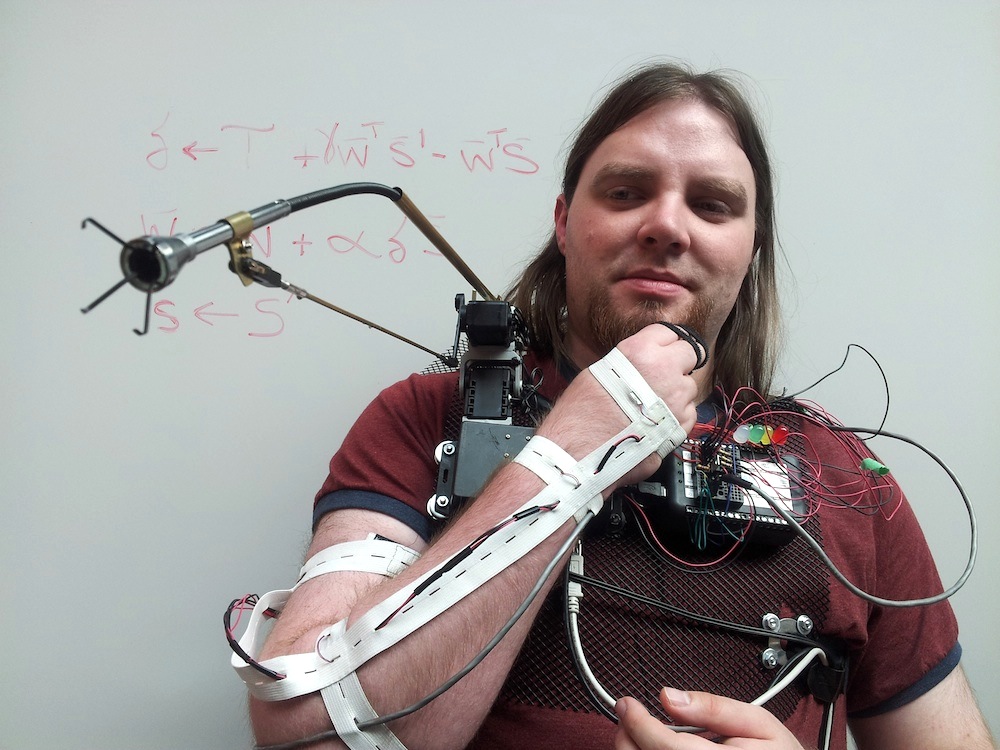}\\
\caption{Wearable robot limb system used in experiments, including a four degree-of-freedom arm, control electronics, and vibrotactile feedback sleeve.}
\label{exarm}
\vspace{-0.6em}
\end{figure}

Artificial intelligence offers a promising solution to the control problems encountered by the users of electromechanical prosthetics\cite{Pilarski2013ram}. Offline machine learning in the form of pattern recognition is for the first time seeing use in commercial prostheses, and is considered to be the state-of-the-art in controlling multiple prosthetic joints \cite{Scheme2011,Micera2010}. 
Real-time machine learning has also recently been used to ease the control burden on a user by learning joint activation sequences as a limb is being used \cite{Pilarski2012biorob,Pilarski2013icorr}; as one example, predictions about a user's control choices have been learned so as to minimize the number of switches between joints, and consequently the time required to perform a task\cite{Edwards2014mec}.

The primary contribution of the present work is to suggest that machine intelligence can be used to enhance not just control---the focus of most prosthesis-related machine intelligence research to date---but also {\em feedback from a prosthesis}.  Feedback is part of a user's biological system, and contains information used in the operation of a natural limb. In the case of a prosthetic limb,  motor awareness and forecasting are now at least partly encoded in the hardware of the prosthesis rather than in a user's biology. Therefore, we may need to provide assistance to the natural system in interfacing with its electronic components. We suggest that a simple artificial intelligence can be used to take the internal state of the assistive device and interpret it in ways the biological system cannot do naturally; the results of this interpretation can be communicated to the user in a variety of ways to improve their control over the device. Thus, using artificial intelligence, we can help create a  forward prediction of an action and communicate it to the user, similar to the operation of the intact biological system.

This work therefore contributes a preliminary exploration of the application of machine-learned predictions, alongside a simple system for communicating those predictions, to assist a user in refining their own forward model of motor actions while using a prosthetic limb. Specifically, temporal-difference learning is used to generate a prediction about the electrical load the servos of a prosthetic limb will experience as they near a potentially dangerous collision with objects in the user's environment. This prediction is communicated to the user through a vibration motor. In this way, we emulate the forward predictive model present in a biological limb's motor function. We expect that, similar to the way that the biological operation of a limb is dependent on its forward copy, the addition of an electronic/computational equivalent during human-robot collaboration will yield control improvements over purely reactive feedback.

\section{METHODS}

\subsection{Robot and Experimental Platform}

\begin{figure}[t]
\centering
\includegraphics[width=0.99\linewidth]{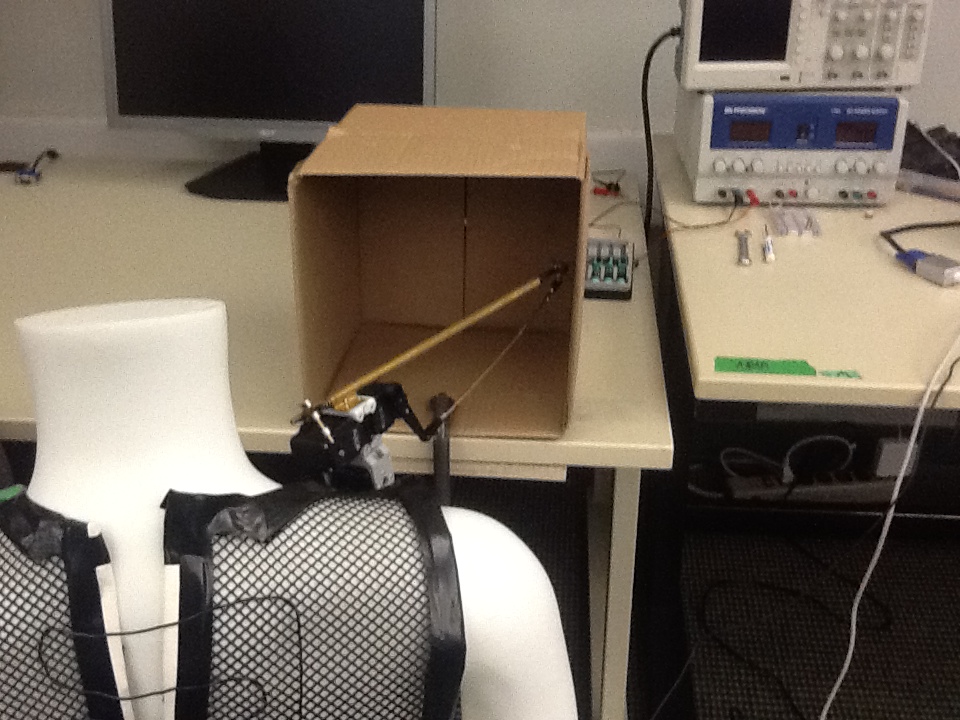}\\
\vspace{0.3em}
\includegraphics[width=0.99\linewidth]{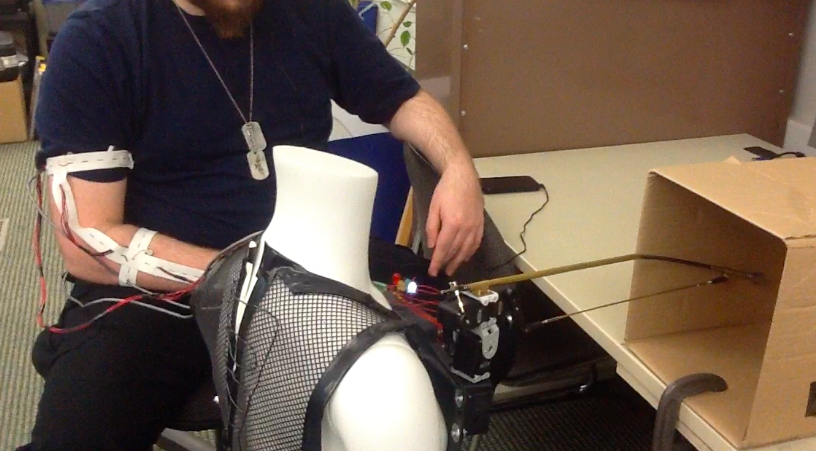}\\
\caption{The experimental setup: a confined workspace, the robotic arm, and an experimental subject with attached vibrotactile feedback sleeve.}
\label{expsetup}
\end{figure}

The experimental platform used in this work was a custom-designed robotic arm called the Extra Robotic \nobreak Manipulator (XRM, Fig.\ref{exarm}), wearable by able-bodied subjects. The arm was designed to model the gross motor functionality of joints in a human arm. It had four controllable actuators: shoulder, elbow, wrist, and hand (AX-12/18+ Dynamixel servo motors). Subjects could use a 2-axis thumb joystick (SparkFun) to switch between and control the motion of the XRM's joints. The joystick was connected to an ADC (DI-149 data acquisition starter kit, DATAQ Instruments), which digitized the 3.3 V signal modified by the user's control of the joystick. The resulting output signal was sent via USB to a computer, which interpreted the signals and sent commands to the robot's servos. The AX-12/18+ servos used in the design of the XRM provided several useful output signals, including their angular position, angular velocity, motor temperature, voltage, and load. To communicate feedback about these sensors to the user, we designed a custom sleeve embedded with four vibration motors (termed {\em tactors}) similar to those used in a cellphone or pager. With the sleeve donned, one tactor each was located over the person's shoulder, elbow, wrist, and hand, as shown in Figs.\ \ref{exarm} and \ref{expsetup}. The platform therefore emulated the capacity for  actuation and vibrotactile feedback found in many common prosthetic devices.

\subsection{Experimental Procedure}

Five subjects were asked to participate in experiments with the XRM, and gave informed consent in accordance with the study's institutional review board approval. Each user wore the sleeve containing the vibration tactors and controlled the left-and-right motion of the robotic arm's shoulder joint using the thumb joystick. The XRM was affixed to a stationary mannequin to ensure each experiment began with the robotic arm at a constant position and to mitigate the effect of a user's trunk movement. Thus, for this initial work, the position and movement of the user was unrelated to the outcome of the experiment. The workspace was a subspace of the arm's total range of motion, bounded by a 27 cm square box that was fastened in place. Prior to each experiment, the end effector was centered with respect to the workspace, perpendicular to the rear wall of the box and equidistant from the left and right walls. Each subject was asked to perform four separate five-minute tasks, structured as follows:

\subsubsection{Training Task} The first task was designed to provide users with practice controlling the XRM. For this training task, the user was asked to move the arm repetitively from one side of the box to the other using the joystick, pausing briefly ($\le$1 second) upon reaching the center of the workspace. At the left and right walls of the box, the user was tasked with pushing the robotic arm against the wall until the arm was fully flexed, causing a temporary increase in the load reported by the servo motor. The user's shoulder vibration tactor was programmed to vibrate at a load threshold of 650 out of a maximum of 1024. This vibration communicated to the user that the load had exceeded the maximum threshold considered safe for the robotic arm, and the arm should be moved away from the wall. In addition to providing each user with practice manipulating the arm, this task produced the source data for prediction learning (described below). 

\subsubsection{No-Feedback Task} Each subject performed the second task without any knowledge of the position of the arm within the workspace other than its starting location. In order to establish a baseline with no visual, auditory, or tactile feedback, subjects were given a blindfold and listened to music through earphones throughout the task. The volume of the music was increased to a comfortable level at which they could not hear the arm tapping the walls of the box. During this task, vibratory feedback about load was also turned off. The instruction given to the user for this task and those that follow was to try to approach the left and right walls in an alternating fashion without making contact.

\subsubsection{Reactive-Feedback Task} The next task was identical to the no-feedback task; however, the participant was provided with reactive vibration feedback when the current load experienced by the robot arm's shoulder servo reached a threshold of more than 420 out of 1024. Thus, the tactor triggered every time the user hit a wall. This task provided an indication of the effectiveness of having reactive tactile feedback only, and specifically examined how well the user could approach each wall without incurring a forceful impact when feedback was delivered at the moment the arm first contacted  the wall.

\subsubsection{Predictive-Feedback Task} For the final task given to participants, users were again blindfolded and sound-isolated. In this case, they were provided with tactile feedback from predictions of the electrical load on the robot's arm servo motor.  Predictions were provided in real time by a machine learning system trained on the interaction data from task 1 as it was being acquired. This prediction learning system is described in the following section. When the load prediction rose above 900, the shoulder tactor was programmed to vibrate. This task was designed to determine how effectively the learning system was able to predict load in advance of hitting each wall, and how communicating this load changed the user's ability to approach the wall without incurring a forceful impact.

All load and prediction thresholds used were determined from the analysis of data prior to experiments. We determined the noise level in each of the tasks; thresholds were then set proportional to this noise level. 

\subsection{Machine Intelligence and Prediction Learning}

The main component of this study is an incremental prediction learner to generate expectations about future impact given learned knowledge about the user's previous motion choices, their outcomes, and the current state of the robot arm. To make predictions about the world, intelligent systems require sensory inputs. These inputs can then be divided into discrete states for increased or decreased resolution. The shoulder joint of the XRM has a rotation range of 300$^\mathsf{o}$. In our protocol, we used angular position of the shoulder joint as a sensory input, divided into 32 distinct states (termed \emph{bins}). These states were motion-dependent; as such, each of the 32 states was further divided into three: one state for clockwise motion, one for counter-clockwise motion, and a third for no motion. The immediate state of the arm was noted in a feature vector (denoted $x$, of length 96) as a single active bit indicating the current position and direction; this feature vector also contained a single active baseline unit.  A weight vector of corresponding length, denoted $w$, was used to store the learned predictions about the interactions between the robot arm and the walls of the workspace. 

The weight vector $w$ was learned from data using standard techniques from temporal-difference learning and recent generalized value function methods, as outlined for the prosthetic setting in Pilarski et al.\cite{Pilarski2013ram} and more generally in Modayil et al. \cite{Modayil2014}. Weights $w$ were updated on each time step according to the temporal difference between the instantaneous load being reported by the servo (denoted $\tau$) and predictions about the immediate and next load readings (the inner products $w_t^\mathsf{T}x_{t}$ and $\gamma w_t^\mathsf{T}x_{t+1}$, respectively, where $\gamma$ is the timescale or level of temporal abstraction for the prediction of interest). The update to the weight vector on each timestep $t$ was done according to:
$$w_{t+1} = w_t + \alpha (\tau_{t+1} + \gamma w_t^\mathsf{T}x_{t+1}- w_t^\mathsf{T}x_t) x_t,$$

where $\alpha$ represents a step-size (learning rate, set to $\alpha=0.1$ in these experiments). The temporal abstraction for predicting the load signal of interest was set to $\gamma=0.92$; this means the prediction learner was acquiring knowledge about the exponentially discounted expectation of the electrical load experienced by the robot's shoulder servo motor over the next $\sim$12 time steps, or 0.6 seconds; the system learned and operated within a control cycle of roughly 20 Hz (50 ms time steps). This knowledge could then be retrieved and used in predictive feedback by reporting the prediction as the inner product $w_t^\mathsf{T}x_{t}$. As noted above, in the predictive feedback task,  vibratory feedback to the user was triggered when the prediction's value exceeded a fixed threshold, indicating an impending collision with the walls of the workspace. 

Learning was only enabled during the training task, such that the system acquired and updated user-specific predictions about servo motor load while each subject was performing their first task. Learning weights were then frozen (i.e., $\alpha=0$) during all remaining tasks, including the predictive feedback task. Learning could in principle continue during all tasks; however, for clear assessment of the principles of interest, our experimental protocol featured defined training and testing periods.

\section{RESULTS}

\begin{figure}[t]
\centering
\includegraphics[width=0.95\linewidth]{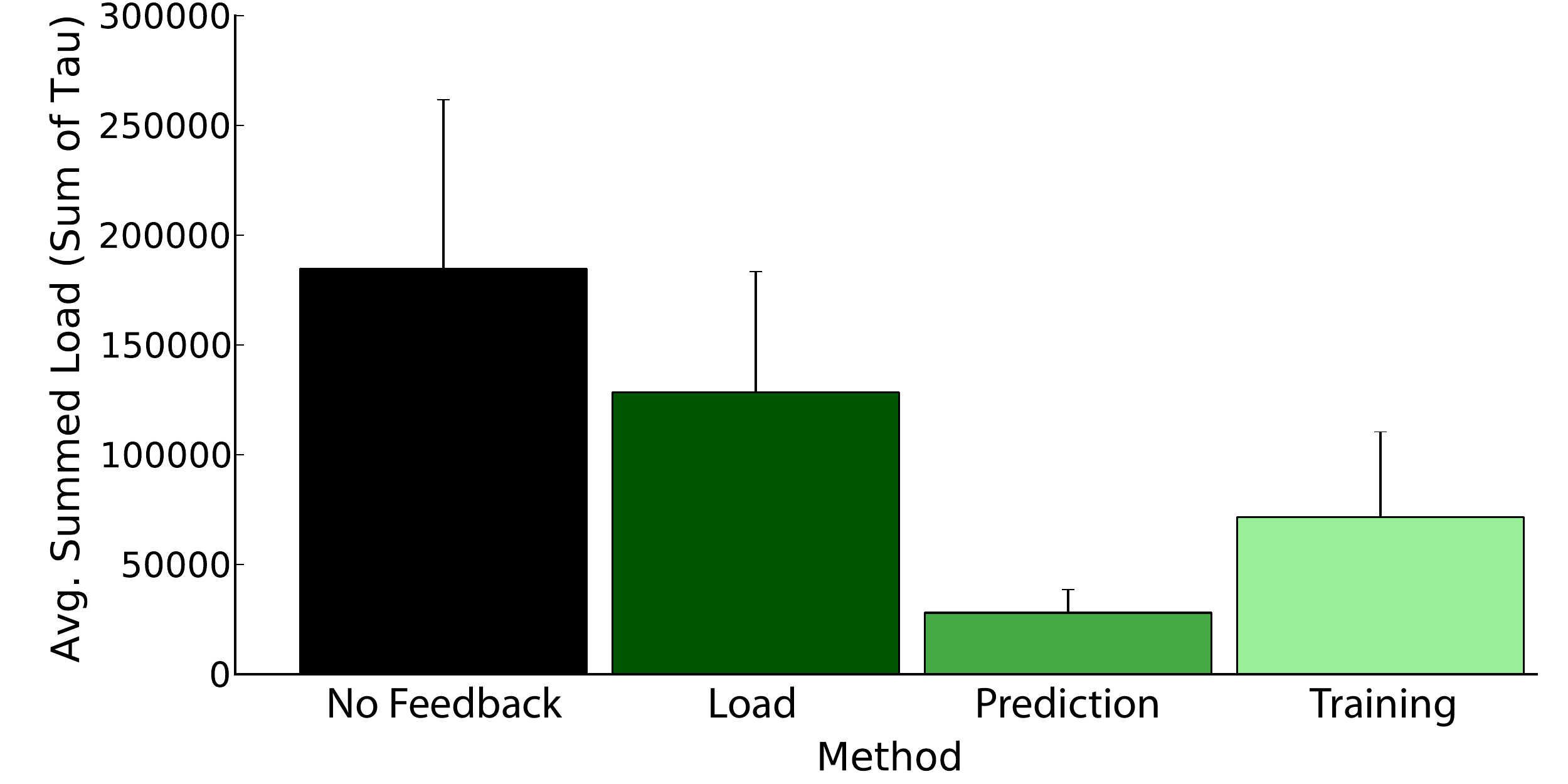}\\
\caption{Key finding: the use of predictive feedback reduced the load (a measure of impact intensity) experienced by the system during use. Aggregate results are shown for all five subjects in terms of the average summed load over the entire length of each subtrial; shown for no feedback (black), load-based feedback (dark green), predictive feedback (light green), and the training data set (lightest green).}
\label{results:keyfinding}
\end{figure}

Giving learned, predictive information as feedback to the user was found to reduce the load experienced by the shoulder actuator of the robot limb when compared to the case where purely reactive feedback was given to the user. As shown in Fig.\ \ref{results:keyfinding}, the total summed load over an entire trial was significantly less with predictive feedback than with reactive feedback. As was expected, load on the robot's motor was visibly higher for all subjects when no visual, auditory, or tactile feedback was provided; with no feedback, different subjects were observed to experience ``control drift'' over the course of the experiment---i.e., to bias their interactions toward one side of the experimental apparatus when provided with no feedback. 

\begin{figure}[t]
\centering
\includegraphics[width=0.95\linewidth]{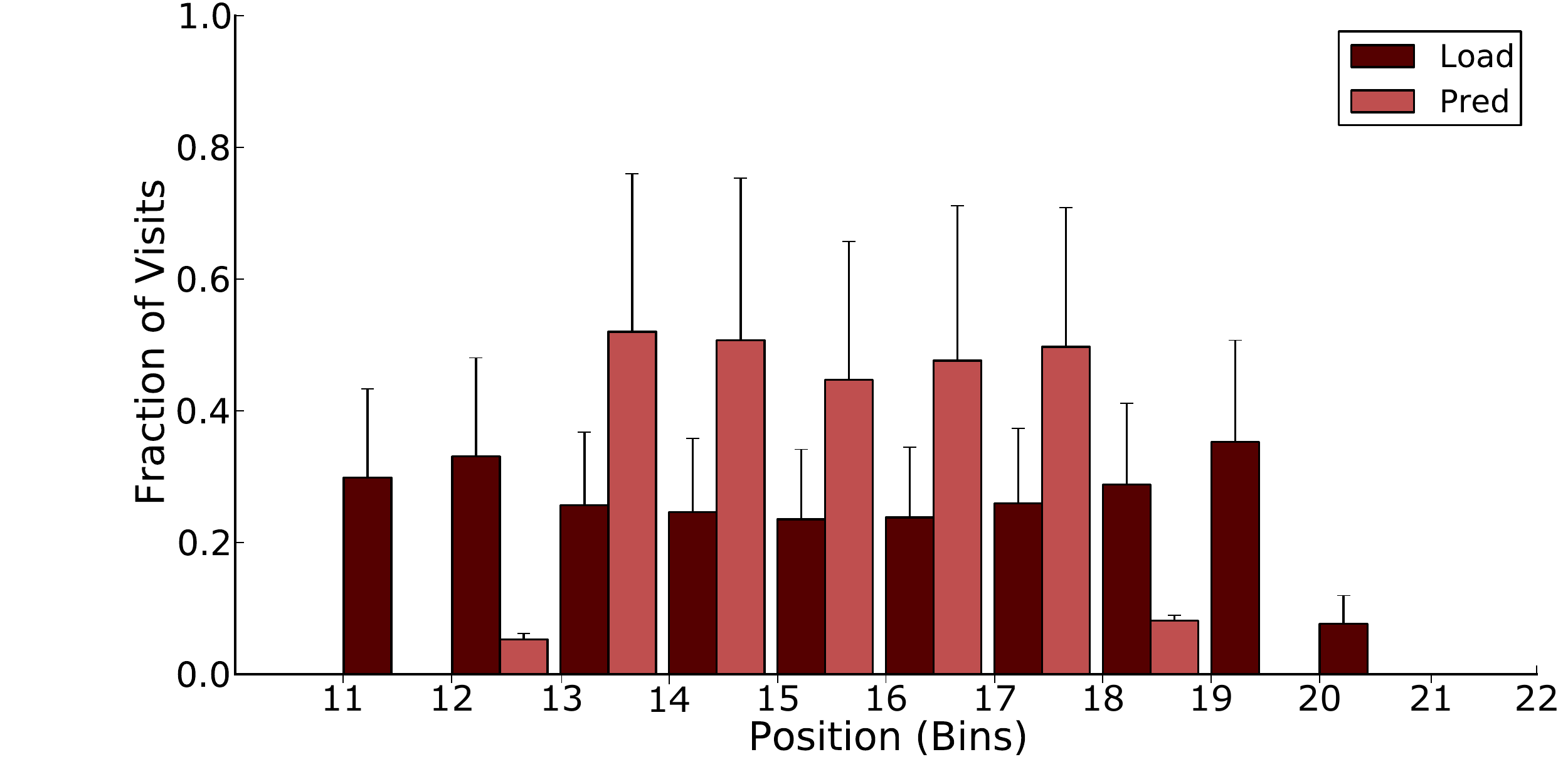}\\
(a)\\
\includegraphics[width=0.95\linewidth]{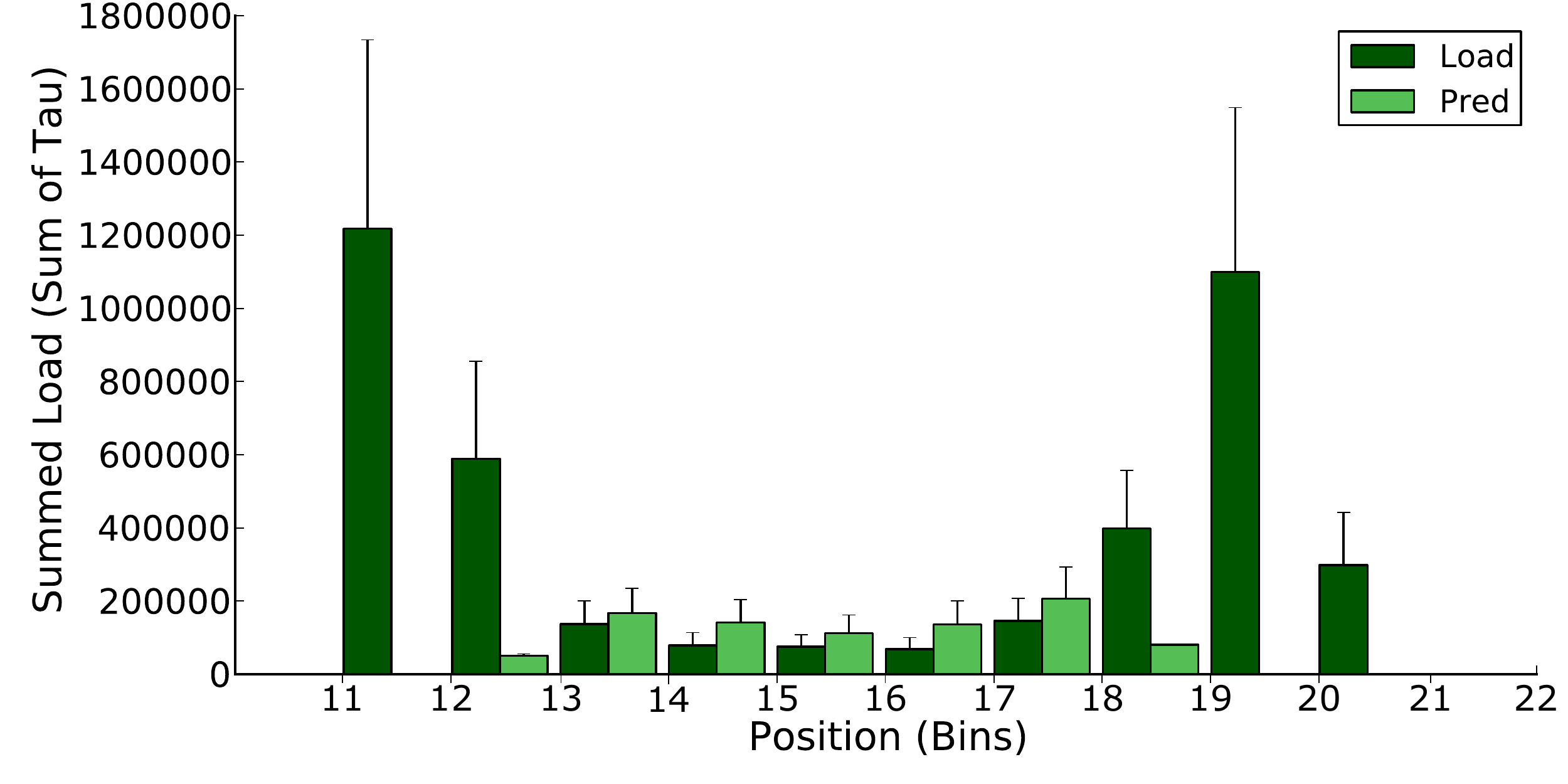}\\
(b)
\caption{Aggregate results for all five subjects showing (a) the frequency of visiting any given servo motor positional bin and (b) the summed load in each bin; comparison shown for load-based feedback (dark red/green) and predictive feedback (light red/green).}
\label{results:aggregate}
\end{figure}

When using reactive feedback, subjects were observed to contact both walls of the workspace with approximately even frequency, with the robot arm deflecting noticeably on both sides due to the contact. When predictive feedback was provided to the user, the robot arm was also observed to approach the two sides of the workspace uniformly, but with much less or no visible deflection to the arm upon contact. These differences in contact with the left and right walls can be seen in Fig.\ \ref{results:aggregate}a. When predictive feedback was used, the shoulder servo of the robot arm spent a smaller fraction of experimental time in its outermost angular positions for this task (few or no visits to bins 11, 12, 18, 19, and 20) than when using purely reactive feedback (approximately uniform visits to all bins). When predictive feedback was in use, the user spent most of their time in the middle of the robot's shoulder actuator's angular range---the area of the workspace away from the walls---as noted by a higher fraction of visits to bins 13--17 in Fig.\ \ref{results:aggregate}a.

Differences between the predictive and reactive feedback case were also observed for all subjects in terms of the bin-by-bin summed load values experienced by the robot's shoulder servo during the trials (Fig.\ \ref{results:aggregate}b).  When using predictive feedback, the shoulder servo of the robot arm experienced significantly less load in the outermost angular positions (little or no summed load in bins 11, 12, 18, 19, and 20) than when using reactive feedback (large summed load in bins 11, 12, 18, 19, and 20). Fig.\ \ref{results:aggregate}b therefore provides a bin-by-bin view on the aggregate results shown in Fig. \ref{results:keyfinding}.

The aggregate results over all five subjects, as shown in Fig.\ \ref{results:aggregate}, were representative of those observed for individual subjects. Load and position plots for a single subject using both reactive and predictive feedback are shown in Fig.~\ref{results:singlesubject}, also plotted alongside their bin-by-bin values during the training case and the no-feedback case. As in aggregate plots, when positional feedback was in use the shoulder servo of the robot arm spent a smaller fraction of experimental time in the outermost angular positions (Fig. \ref{results:singlesubject}a) and recorded no electrical load significantly above that originating from robot arm motion through free space (Fig. \ref{results:singlesubject}b). 

\section{DISCUSSION}

\begin{figure}[t]
\centering
\includegraphics[width=0.95\linewidth]{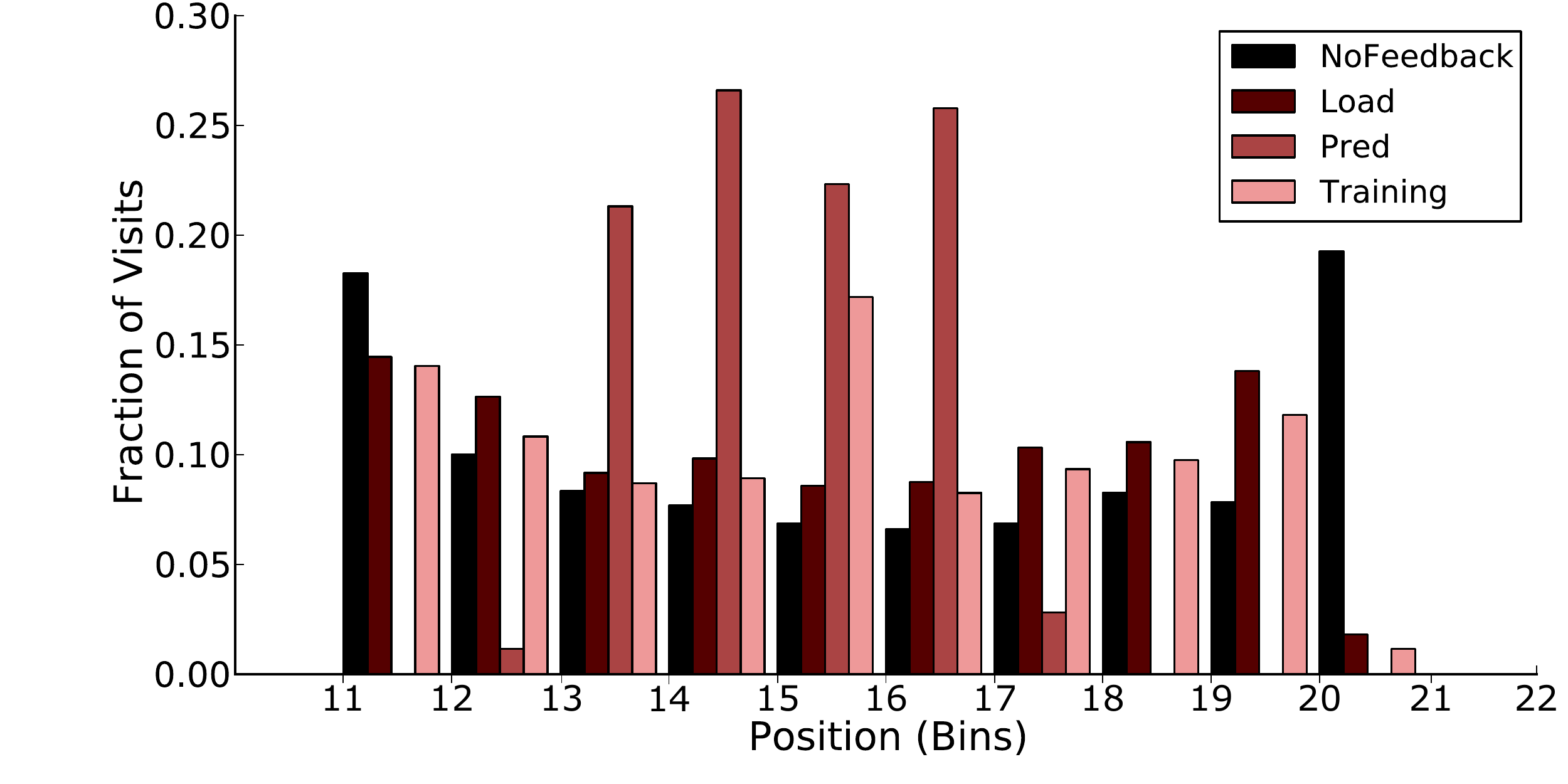}\\
(a)\\
\includegraphics[width=0.95\linewidth]{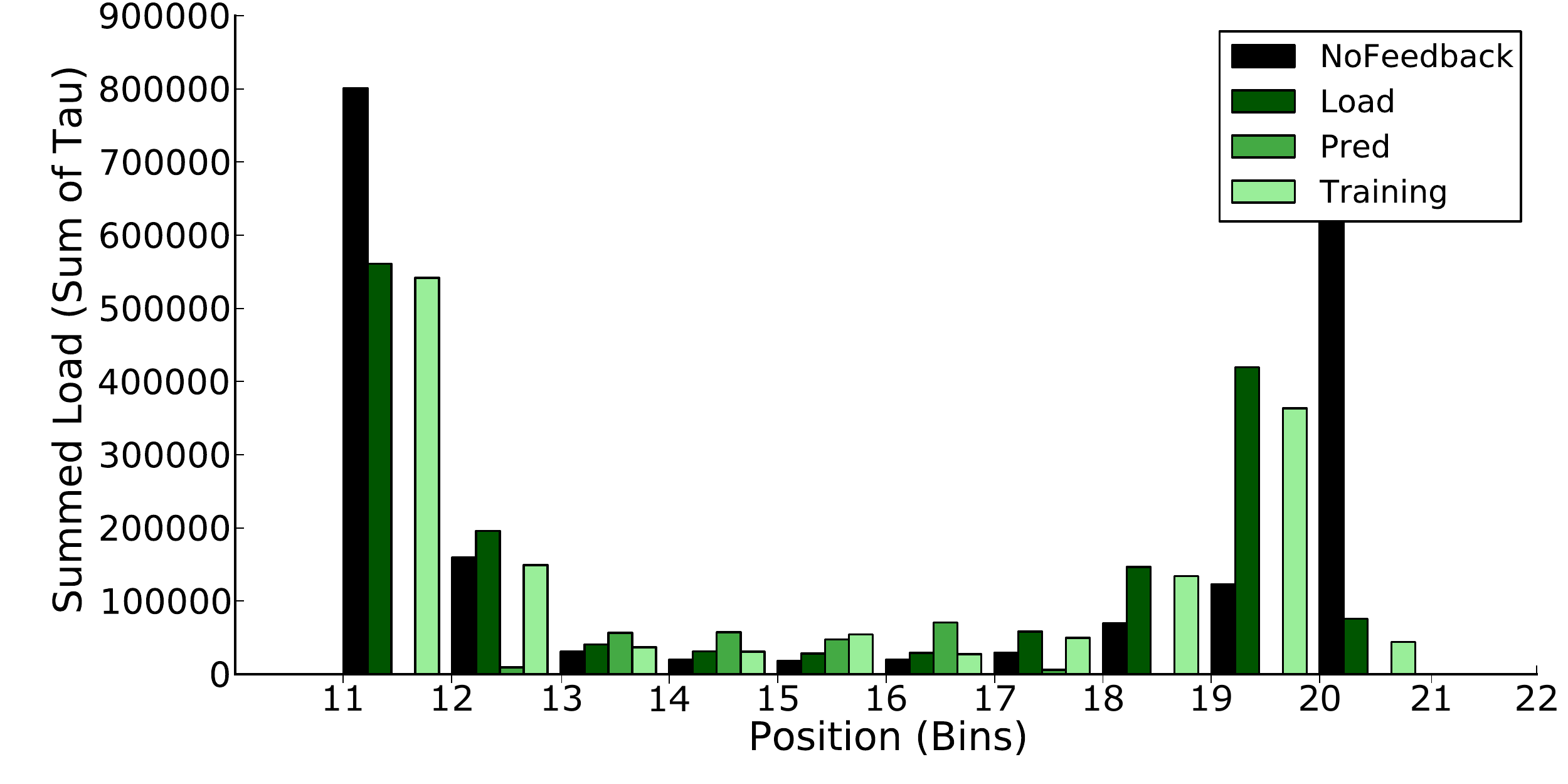}\\
(b)
\caption{Results for a single subject showing (a) the frequency of visiting any given servo motor positional  bin and (b) the summed load in each bin; shown for no feedback (black), load-based feedback (medium), predictive feedback (light), and the training data (lightest).}
\label{results:singlesubject}
\vspace{-1em}
\end{figure}

Feedback is an important aspect of skilled control. As noted above, we defined the control of the robotic device to be successful and skilled if the load experienced by the device while moving near the border of the work area is low---the task objective given to our subjects during testing was to closely approach but not impact the walls of the workspace. With different forms of feedback or different settings, we expected a subject might never get near the wall (overly sensitive predictions, thresholds, or too much temporal extension), that they might do so with high variability (as when operating with minimal feedback), or that they might impact the wall consistently but forcefully if feedback comes too late (e.g., with overly delayed or reactive feedback). Our observations support these expectations. 

Figure \ref{results:keyfinding} demonstrates the effect that different types of feedback had on skilled control of the robotic device. In the no-feedback case, the load experienced by the device was high and variable. Further, the observed bias our users demonstrated to one side or another was well out of line with the desired operational outcome. We expect this level of unskilled control when a user cannot see, hear, or feel the device being operated. A logical solution to this problem is to provide feedback to the user when the load increases beyond a specified point, indicating that a collision with the border of the workspace has occurred. This is shown in Fig.\ \ref{results:keyfinding} to improve upon the case where the user has no feedback, though the cumulative load is still high. No matter how sensitive the threshold is to initiate the reactive feedback, the user must still perceive the feedback and act in an appropriate, timely way; load is inevitable, since it is already occurring, but can potentially be reduced via fast human reaction time---subject-specific reaction time is one possible source of the variance in Fig.\ \ref{results:keyfinding}. By far the most successful operation in terms of both load and adherence to the boundaries of the workspace resulted from the use of predictive feedback. With this anticipatory form of prompting, the load was significantly lower and less varied than in any other trial, indicating successful task performance with the device despite the user's inability to see or hear their status. 

	A more detailed indication of the motion of the device is illustrated in Fig.\ \ref{results:aggregate}. In particular, Fig.\ \ref{results:aggregate} highlights differences in predictive and reactive feedback in the area of travel between the borders of the workspace (between bins 13 and 17). In this region, the bin-by-bin cumulative load (Fig.~\ref{results:aggregate}b), as well as the fraction of visits (Fig.~\ref{results:aggregate}a), are both shown to be higher than the same measures using reactive load feedback. This is an indication of successful operation, as it shows that the device spent more time transitioning from border to border rather than impacting them. This observation is further supported by the significant differences that can be seen between the two types of feedback at and beyond bins 12 and 18---the borders of the workspace. With predictive feedback, the user moved into the borders of the workspace less frequently (Fig.\ \ref{results:aggregate}a). In doing so, the system experienced significantly less (or no) load in bins 11--12 and 18--20 (Fig.~\ref{results:aggregate}b), indicating less time spent under impact conditions or flexing of the physical device. With reactive load feedback, the system experienced significantly greater load at the borders than it did while transitioning between the borders. As noted above, early stopping might be expected if the predictive feedback was overly sensitive or too anticipatory. With predictive feedback, the fraction of visits for each bin was roughly uniform up until the workspace boundaries (Fig.~\ref{results:aggregate}a), indicating that the user was not stopping too far in advance of the wall. From the single subject results in Fig.~\ref{results:singlesubject}, we can draw much the same conclusions as we do from the aggregate results in Fig.\ \ref{results:aggregate}. 

\subsection{Tuning, Training, and Adaptability}

With predictive feedback and the settings described above, we qualitatively observed subjects stopping the robot arm's motion such that it made only light, unloaded contact with the wall. This level of contact could be modified by varying the learning parameters of the artificial intelligence, and parameters could be adjusted in a number of ways to achieve a number of outcomes. There is no one ``correct'' setting for sensitivity; instead, there are a number of possibilities for how the device can assist the user in achieving their objectives. Learning parameters could be tuned to provide feedback behaviour that duplicates that of the reactive feedback case. The converse is not true---reactive feedback is not capable of providing preemptive feedback about future events. Also, when using predictive feedback, we observed that the threshold for indicating an impending load could be made less sensitive than the equivalent reactive load without incurring false positives. As the learned predictions are mathematical expectations conditioned on servo position, they are not affected by spurious load variance or noise due to motion, as would be a purely reactive approach. 

For clarity of assessment, the artificial intelligence system in this preliminary study only acquired and updated its predictive knowledge during a defined training period. In any machine learning setting with a fixed training period, variability in training can noticeably affect learning system performance, but should not affect fixed or reactive approaches. As shown in Fig.\ \ref{results:singlesubject}, we observed a subject-specific difference to the visits and load in bin 17 while using predictive feedback---differences to the training of the learning system or a slight shift in the experimental setup may have resulted in an earlier feedback prompt to this user in terms of absolute servo-motor position on one side of the workspace. Omissions during training or changes to the domain of use may be corrected or updated through the use of continuing or ongoing machine learning. This has been suggested in prior work \cite{Pilarski2013ram}, and is a natural way to robustly extend the present study. As learning is already done in a per-time-step, incremental way during  training, there are no technical or algorithmic barriers to continuing the learning of feedback-related predictions during operational use. While many offline or batch prediction learning methods could potentially be used to generate expectations for use in feedback (e.g., the work of  Pulliam et al.\cite{Pulliam2011}), the continuing and computationally inexpensive nature of our chosen learning approach makes it  well suited to use in a prosthetic environment \cite{Pilarski2013ram}. Our prediction learning approach is suitable for subject-specific, task-specific learning with no requirement for a priori domain knowledge; it is also well suited for adapting to ongoing changes in a task or a user's behaviour during persistent, real-time use.

\subsection{Feedback Modalities}

	As noted above, much work is being done to restore missing feedback to prosthesis users \cite{Antfolk2013,Hebert2014}. Focus has been placed on restoring touch, including sensations such as pressure, texture, temperature, and even pain. A large body of this research has explored feedback using sensory substitution, wherein one sensation is replaced with another different sensation that the user must be trained to skillfully interpret; use of this approach is largely due to the physiological constraints of prosthetic human-machine interaction \cite{Antfolk2013}. Modality-matched feedback is also receiving growing attention; in matched feedback, sensations are restored either invasively or non-invasively to the natural or proxy locations that convey sensations of the lost or damaged biological system as closely as possible \cite{Antfolk2013,Hebert2014}. 
	
Our present study can be thought of as a form of substitution feedback---predictions about the electrical current drawn by the device during operation (perhaps thought of as the device's ``pain'' or motor fatigue) are communicated to the user via a vibratory buzzing sensation in order to prompt the user to take action to prevent it. This buzzing is not a natural sensation, and it is not communicated at an equivalent natural location on the user's body.  What separates this choice from the usual form of sensory substitution is that fact that the information being transmitted from the user to the device is not a biological sensation---it is specific to the internal hardware of the device and encodes a prediction about future changes to that hardware. While communicating these anticipations is helpful to the successful operation of the device, it is not a natural thing for the user to feel; as with most substitution feedback, it takes training to interpret such a sensation (as noted in Hebert et al.\ \cite{Hebert2014}). This training need was perhaps minimized for our test subjects because of the precedent in modern society to interpret the vibration of personal device as a prompt to act (e.g., cellphone vibration in response to a new text message).

However, our work should not be thought of solely in terms of sensory substitution. Our study is intended to be a small window into a larger area for research: the use of machine intelligence as a method for filtering, selecting, and communicating salient information about the internal state of a complex device. This communication can be thought of as a form of {\em transparency}, as used by Thomaz and Breazeal \cite{Thomaz2008}. Communication of such non-biological knowledge to the device's user---e.g., prompts regarding a device's internal state, decisions, and anticipatory knowledge---promises to streamline human-machine interaction in many domains, and should be equally suited to  feedback via both sensory substitution and modality-matched percepts.

\section{Future Work}

The results presented in this work are preliminary, and there is much room for further study in this area. The incremental learning algorithm used in this experiment was effective but monolithic. If a control-learning system were used in conjunction with the present prediction-learning algorithm, it may be possible for a device to adapt the timing and magnitude of its feedback to better suit its domain of use. For instance, the feedback threshold or level of temporal abstraction $\gamma$ could be tuned on the basis of reward-like signals of approval or disapproval delivered by the user, using techniques from related work on the human training of machine learners \cite{Thomaz2008,Pilarski2011icorr,Knox2013iui}; predictive load information could be communicated at distances from the collision which have been learned to be appropriate for a specific user and their task preferences. Further, as artificial intelligence use in prosthetic limbs becomes more prevalent, communicating the actions that the artificial intelligence has learned to take to the user may help allow more control to pass to the prosthetic---the case of shared control and sliding-scale autonomy. Transparent communication between the operator and their device could be the keystone which allows an intelligent prosthetic and a human user to co-operate, combine processing power, and more effectively restore lost function.

\section{CONCLUSIONS}

Feedback is important to prosthetic limb control. While machine intelligence has been used to improve the interpretation of control signals given to a limb from the user, its use in modulating feedback is often overlooked. This article contributed an initial look at one way predictions and machine learning may be used in feedback to close the loop between a human and their artificial limb. To our knowledge, this is the first study investigating the use of machine intelligence in prosthetic feedback. 

When compared to strictly communicating momentary electrical load to the user, communicating a machine-learned forecast of the same load was found to decrease the impacts experienced by the robot limb, and to increase the ability of our subjects to position the robot limb despite the absence of all other feedback. The increase in precision in terms of both position and load over the no-feedback case was dramatic, and the increase over purely reactive feedback was significant. Though preliminary, these results promise two related outcomes for the user of a prosthetic limb. First, we expect that increased communication from the device about its internal state and setting of use may allow the user more personalized and more trustworthy options for control. Over the long term, predictive feedback could therefore lead to greater acceptance and assimilation of the device as part of the user. Further, by creating a computational predictive forward copy of an action and communicating it to the user, operating an assistive device may become more precise. These expectations remain to be verified during the use of predictive feedback in real-life functional tasks.

\addtolength{\textheight}{-12cm}   




\section*{ACKNOWLEDGMENT}

The authors thank Joseph Modayil, Richard Sutton, and Jim Parker for insights and suggestions relating to this work.



\end{document}